\documentclass[10pt,twocolumn,letterpaper]{article}

\usepackage{cvpr}              % To produce the CAMERA-READY version
\usepackage[mode=buildnew]{standalone}
\usepackage{booktabs}
\usepackage{amsmath,amsthm,amssymb,amsfonts}
\usepackage{multicol}
\usepackage{multirow}
\usepackage{tikz}
\usepackage{forest}
\usepackage{xcolor}
\usetikzlibrary{shadows}
\usepackage{colortbl}
\definecolor{tu0}{rgb}{0.7451, 0.1176, 0.2353}

\definecolor{tu1}{rgb}{1.0000, 0.8039, 0.0000}
\definecolor{tu11}{rgb}{1.0000, 0.8627, 0.3020}
\definecolor{tu12}{rgb}{1.0000, 0.9020, 0.4980}
\definecolor{tu13}{rgb}{1.0000, 0.9412, 0.6980}
\definecolor{tu14}{rgb}{1.0000, 0.9608, 0.8000}

\definecolor{tu2}{rgb}{0.9804, 0.4314, 0.0000}
\definecolor{tu21}{rgb}{0.9882, 0.6039, 0.3020}
\definecolor{tu22}{rgb}{0.9882, 0.7137, 0.4980}
\definecolor{cscolor}{rgb}{0.9882, 0.7137, 0.4980}
\definecolor{tu23}{rgb}{0.9922, 0.8275, 0.6980}
\definecolor{tu24}{rgb}{0.9961, 0.8863, 0.8000}

\definecolor{tu3}{rgb}{0.6902, 0.0000, 0.2745}
\definecolor{tu31}{rgb}{0.7529, 0.2000, 0.4196}
\definecolor{tu32}{rgb}{0.8431, 0.4980, 0.6353}
\definecolor{mvcolor}{rgb}{0.8431, 0.4980, 0.6353}
\definecolor{tu33}{rgb}{0.9216, 0.7490, 0.8196}
\definecolor{tu34}{rgb}{0.9529, 0.8510, 0.8902}

\definecolor{tu4}{rgb}{0.4863, 0.8039, 0.9020}
\definecolor{tu41}{rgb}{0.6431, 0.8627, 0.9333}
\definecolor{tu42}{rgb}{0.7412, 0.9020, 0.9490}
\definecolor{bddcolor}{rgb}{0.7412, 0.9020, 0.9490}
\definecolor{tu43}{rgb}{0.8431, 0.9412, 0.9686}
\definecolor{tu44}{rgb}{0.8980, 0.9608, 0.9804}

\definecolor{tu5}{rgb}{0.0000, 0.5020, 0.7059}
\definecolor{tu51}{rgb}{0.3020, 0.6510, 0.7961}
\definecolor{tu52}{rgb}{0.5490, 0.7765, 0.8667}
\definecolor{tu53}{rgb}{0.7490, 0.8745, 0.9255}
\definecolor{tu54}{rgb}{0.8510, 0.9255, 0.9569}

\definecolor{tu6}{rgb}{0.0000, 0.3255, 0.4549}
\definecolor{tu61}{rgb}{0.2510, 0.4941, 0.5922}
\definecolor{tu62}{rgb}{0.5490, 0.6941, 0.7529}
\definecolor{tu63}{rgb}{0.7490, 0.8314, 0.8627}
\definecolor{tu64}{rgb}{0.8510, 0.8980, 0.9176}

\definecolor{tu7}{rgb}{0.0314, 0.0314, 0.0314}
\definecolor{tu71}{rgb}{0.3725, 0.3725, 0.3725}
\definecolor{tu72}{rgb}{0.5882, 0.5882, 0.5882}
\definecolor{tu73}{rgb}{0.7529, 0.7529, 0.7529}
\definecolor{tu74}{rgb}{0.8667, 0.8667, 0.8667}

\definecolor{tu8}{rgb}{0.7765, 0.9333, 0.0000}
\definecolor{tu81}{rgb}{0.8431, 0.9529, 0.3020}
\definecolor{tu82}{rgb}{0.8863, 0.9647, 0.4980}
\definecolor{acdccolor}{rgb}{0.8863, 0.9647, 0.4980}
\definecolor{tu83}{rgb}{0.9333, 0.9804, 0.6980}
\definecolor{tu84}{rgb}{0.9569, 0.9882, 0.8000}

\definecolor{tu9}{rgb}{0.5373, 0.6431, 0.0000}
\definecolor{tu91}{rgb}{0.6784, 0.7490, 0.3020}
\definecolor{tu92}{rgb}{0.7686, 0.8196, 0.4980}
\definecolor{tu93}{rgb}{0.8588, 0.8941, 0.6980}
\definecolor{tu94}{rgb}{0.9059, 0.9294, 0.8000}

\definecolor{tu10}{rgb}{0.0000, 0.4431, 0.3373}
\definecolor{tu101}{rgb}{0.3020, 0.6118, 0.5373}
\definecolor{tu102}{rgb}{0.5490, 0.7490, 0.7020}
\definecolor{tu103}{rgb}{0.7490, 0.8588, 0.8353}
\definecolor{tu104}{rgb}{0.8549, 0.9176, 0.9059}

\definecolor{tu110}{rgb}{0.8000, 0.0000, 0.6000}
\definecolor{tu111}{rgb}{0.8706, 0.3490, 0.7412}
\definecolor{tu112}{rgb}{0.9216, 0.6000, 0.8392}
\definecolor{gta5color}{rgb}{0.9216, 0.6000, 0.8392}
\definecolor{tu113}{rgb}{0.9608, 0.8000, 0.9216}
\definecolor{tu114}{rgb}{0.9804, 0.8980, 0.9608}

\definecolor{tu120}{rgb}{0.4627, 0.0000, 0.4627}
\definecolor{tu121}{rgb}{0.5961, 0.2510, 0.5961}
\definecolor{tu122}{rgb}{0.7294, 0.4980, 0.7294}
\definecolor{tu123}{rgb}{0.8392, 0.6980, 0.8392}
\definecolor{tu124}{rgb}{0.9216, 0.8510, 0.9216}

\definecolor{tu130}{rgb}{0.4627, 0.0000, 0.3294}
\definecolor{tu131}{rgb}{0.6118, 0.3020, 0.5333}
\definecolor{tu132}{rgb}{0.7569, 0.5490, 0.6980}
\definecolor{synthiacolor}{rgb}{0.7569, 0.5490, 0.6980}
\definecolor{tu133}{rgb}{0.8667, 0.7490, 0.8314}
\definecolor{tu134}{rgb}{0.9216, 0.8510, 0.9020}

\definecolor{con_blue}{HTML}{004488}
\definecolor{con_yellow}{HTML}{DDAA33}
\definecolor{con_red}{HTML}{BB5566}
\definecolor{con_black}{HTML}{000000}
\definecolor{con_gray}{HTML}{AAAAAA}

\definecolor{vib_orange}{HTML}{EE7733}
\definecolor{vib_darkblue}{HTML}{0077BB}
\definecolor{vib_lightblue}{HTML}{33BBEE}
\definecolor{vib_magenta}{HTML}{EE3377}
\definecolor{vib_red}{HTML}{CC3311}
\definecolor{vib_green}{HTML}{009988}
\definecolor{vib_gray}{HTML}{BBBBBB}
\definecolor{vib_black}{HTML}{000000}

\newcommand{\src}[0]{{\mathcal{D}^\mathrm{S}}}

\newcommand{\network}[1]{{#1}}

\newcommand{\csval}[0]{{\mathcal{D}^\mathrm{CS}_\mathrm{val}}}

\newcommand{\bddval}[0]{{\mathcal{D}^\mathrm{BDD}_\mathrm{val}}}

\newcommand{\mvval}[0]{{\mathcal{D}^\mathrm{MV}_\mathrm{val}}}

\definecolor{textcolor}{RGB}{70,170,34}

\newcommand{\putindex}[3]{\vtop{\hbox{\hspace{#3} $#1$}
            \hbox{\raise 6mm \hbox{$\scriptscriptstyle #2$}}}}

\newcommand{\gradx}[0]{\vtop{\hbox{\rm grad}
            \hbox{\raise 2.5mm \hbox{\rm \hspace{2mm} \footnotesize x}}}}

\newcommand{\grady}[0]{\vtop{\hbox{\rm grad}
            \hbox{\raise 2.5mm \hbox{\rm \hspace{2mm} \footnotesize y}}}}

\newcommand{\grad}[1]{\vtop{\hbox{\rm grad}
            \hbox{\raise 2.5mm \hbox{#1}}}}

\newcommand{\btb}{     \begin{tabbing}             }
\newcommand{\bte}{     \end{tabbing}               }

\definecolor{cvprblue}{rgb}{0.21,0.49,0.74}
\usepackage[pagebackref,breaklinks,colorlinks,allcolors=cvprblue]{hyperref}

\def\paperID{12} % *** Enter the Paper ID here
\def\confName{CVPR}
\def\confYear{2025}
\newcolumntype{g}{>{\columncolor{tu73}}c}
\newcolumntype{h}{>{\columncolor{tu74}}c}

\title{Domain Generalization for Semantic Segmentation: A Survey}

\author{Manuel Schwonberg$^{1,2}$ , Hanno Gottschalk$^{1}$ \\
$^{1}$TU Berlin, Mathematical Modeling of Industrial Life Cycles Group\\
$^{2}$CARIAD SE
 \\
{\tt\small schwonberg@campus.tu-berlin.de, gottschalk@math.tu-berlin.de}
}

\begin{document}
\flushbottom 
\maketitle
\begin{abstract}
    The generalization of deep neural networks to unknown domains is a major challenge despite their tremendous progress in recent years. For this reason, the dynamic area of domain generalization (DG) has emerged. In contrast to unsupervised domain adaptation, there is no access to or knowledge about the target domains, and DG methods aim to generalize across multiple different unseen target domains. Domain generalization is particularly relevant for the task semantic segmentation which is used in several areas such as biomedicine or automated driving. This survey provides a comprehensive overview of the rapidly evolving topic of domain generalized semantic segmentation. We cluster and review existing approaches and identify the paradigm shift towards foundation-model-based domain generalization. Finally, we provide an extensive performance comparison of all approaches, which highlights the significant influence of foundation models on domain generalization. This survey seeks to advance domain generalization research and inspire scientists to explore new research directions.  
\end{abstract}
\section{Introduction}
Over the last decade, deep neural networks (DNNs) have paved the way for major progress in the field of computer vision \cite{kirillov2023segment, radford2021learning, dosovitskiy2020image, krizhevsky2012imagenet}. That includes dense perception tasks such as semantic segmentation and object detection, which are relevant for several different applications such as automated driving. However, domain shifts, i.e. when the training and inference domain differs, challenge DNNs and cause significant performance decreases when applied in the target domain. In particular, in the field of automated driving, domain shifts are a huge challenge. Several domain shifts, like the synthetic-to-real and real-to-real shifts, are major obstacles to the perception of automated vehicles in the real world. Therefore, several research fields have emerged to mitigate the performance decrease caused by domain shifts. Unsupervised domain adaptation (UDA) is very popular, and UDA for semantic segmentation has gained a lot of attention in research, with a large diversity of approaches \cite{Schwonberg2023Survey, hoyer2022daformer, xie2022sepico, tsai2018learning, hoffman2018cycada}. UDA assumes that unlabeled target data is given and aims for the adaptation to this particular domain. However, unlabeled target data is not always available, and adaptation to a particular, single domain can be hard because of the large number of different domains. In contrast, domain generalization (DG) approaches overcome these flaws by only accessing the source domain and aiming to generalize to multiple unseen target domains. In recent years, a highly relevant and dynamic research field of DG approaches has emerged with a wide variety of approaches. This necessitates a survey on this topic to first get an overview about the variety of existing approaches and second identify important research trends and future research. Although there are general surveys on domain generalization \cite{wang2022generalizing, zhou2022domain}, these works do not provide an in-depth review of DG for semantic segmentation. Only the survey by Rafi \etal \cite{rafi2024domain} specifically addresses domain generalization for segmentation, but in our survey we cover a larger number of approaches, include foundation-model-based DG approaches, and provide a more fine-grained taxonomy. Our survey focuses on synthetic-to-real domain generalization because of the large number of approaches and the uniform benchmarking. Overall, the contributions of this survey are as follows:
\begin{itemize}
    \item we review and cluster over 40 synthetic-to-real domain generalization approaches for DG making it the most extensive survey of DG for semantic segmentation so far
    \item we identify and analyze the paradigm shift from classic DG approaches towards foundation-model-based DG approaches
    \item we provide an extensive performance comparison of all reviewed approaches which reveals differences between source datasets and underlines the influence of foundation models 
\end{itemize}
Our survey is structured as follows. First, we will introduce and explain our taxonomy which highlights the paradigm shift towards foundation models. Subsequently, we will review and analyze each of the categories of the taxonomy in detail and conclude with an extensive performance comparison of the approaches.
  
\begin{figure*}
    \centering
        \tikzset{
        my node/.style={
            %draw=gray,
            % inner color=gray!5,
            % outer color=gray!10,
            thick,
            minimum width=1cm,
%           rounded corners=3,
            %text height=1.5ex,
            %text depth=0ex,
            font=\sffamily,
            %drop shadow,
        }
    }
    \begin{forest}
        for tree={%
            my node,
            l sep+=5pt,
            grow'=south,
            edge={gray, thick},
            parent anchor=south,
            child anchor=north,
            align=center,
            if n children=0{tier=last}{},
            edge path={
                \noexpand\path [draw, \forestoption{edge}] (!u.parent anchor) +(0, 0pt) -- +(0pt,-0pt)  -- (.child anchor)\forestoption{edge label};
            },
            if={isodd(n_children())}{
                for children={
                    if={equal(n,(n_children("!u")+1)/2)}{calign with current}{}
                }
            }{}
        }
        [\Large \textbf{Domain Generalization}, fill=gray!60
        [\large {Foundation-Model-based}, fill=tu82 [Feature Space, fill=tu42 [Vision-Text\\Alignment, fill=tu42!75[tqdm \cite{pak2025textual} \\ MFuser \cite{mfuser}]][Efficient \\Fine-Tuning, fill=tu42!75[Rein \cite{wei2024stronger}\\FADA \cite{bilearning}\\SoRA \cite{yun2024sora}\\ SET \cite{yi2024learning}\\ VLTSeg \cite{hummer2024strong} \\ FisherTune \cite{zhao2025fishertune} \\ MFuser \cite{mfuser}]]]
        ]
        [\large Classic Approaches, fill=tu92 [Output Space, fill=tu22
        [Consistency, fill=tu22!60[SiamDoGe \cite{wu2022siamdoge}\\ GTR \cite{peng2021global}\\ SHADE \cite{zhao2022style}\\WildNet \cite{Lee2022wildnet}\\SRMA \cite{jiao2024semantic}\\ CSDS \cite{liao2024class}]]
        ] 
        [Feature Space, fill=tu42 
        [Contrastive\\ Learning, fill=tu42!75[SPC \cite{huang2023style}\\ BlindNet \cite{ahn2024style}\\DPCL \cite{yang2023generalized}\\ CDPCL \cite{liao2023calibration}\\PtM \cite{kim2022pin}\\ WildNet \cite{Lee2022wildnet}]]
        [Feature\\ Augmentation, fill=tu42!75[SiamDoGe \cite{wu2022siamdoge}\\MRFP \cite{udupa2024mrfp}\\ SPC \cite{huang2023style}\\FGSS \cite{zhang2023fine}\\PtM \cite{kim2022pin}\\WEDGE \cite{kim2021wedge}\\ SHADE \cite{zhao2022style}\\WildNet \cite{Lee2022wildnet}\\RICA \cite{sun2023augment}\\SRMA \cite{jiao2024semantic}\\FAmix \cite{fahes2024simple}]]
        [Feature\\ Regularization, fill=tu42!75[DIRL \cite{xu2022dirl}\\BlindNet \cite{ahn2024style}\\IBN-Net \cite{pan2018two}\\ISW \cite{choi2021robustnet}\\ SW \cite{pan2019switchable}\\SAN+SAW \cite{Peng2022semanticaware}\\ CFC \cite{li2022learning}\\SRMA \cite{jiao2024semantic}\\ ReVT \cite{Termoehlen2023}\\ SHADE \cite{zhao2022style}\\ TLDR \cite{kim2023texture}]]
        [Feature \\Selection, fill=tu42!75 [DIRL \cite{xu2022dirl}\\FSDR \cite{huang2021fsdr}\\ISW \cite{choi2021robustnet}\\ IBAFormer \cite{sun2023ibaformer}\\CMFormer \cite{bi2024learning}]]
        ]
        [Input Space, fill=tu102
        [Image\\ Augmentation, fill=tu102!60[DPCL \cite{yang2023generalized}\\ DRPC \cite{yue2019domain}\\ AdvStyle \cite{zhong2022adversarial}\\FSDR \cite{huang2021fsdr}\\GTR \cite{peng2021global}\\ TLDR \cite{kim2023texture}\\RICA \cite{sun2023augment}\\PASTA \cite{2023iccv_PASTA}\\SIL \cite{zhang2023learning}\\MoDify \cite{jiang2023domain}\\CSDS \cite{liao2024class}\\ DGinStyle \cite{jia2023dginstyle}\\CLOUDS \cite{benigmim2024collaborating}\\ IBAFormer \cite{sun2023ibaformer} \\ DIDEX \cite{niemeijer2024generalization}]]
        [Extra (Real) \\ Data, fill=tu102!60[FSDR \cite{huang2021fsdr}\\ WEDGE \cite{kim2021wedge}\\DRPC \cite{yue2019domain} \\ GTR \cite{peng2021global}\\ WildNet \cite{Lee2022wildnet} \\ TLDR \cite{kim2023texture}\\ CSDS \cite{liao2024class}]]]
        ]]
    \end{forest}
    \caption[Taxonomy of Domain Generalization Approaches]{\textbf{Taxonomy of Domain Generalization Approaches for Semantic Segmentation}}
    \label{fig:dg_taxonomy}
    %\vspace{-3.5mm}
\end{figure*}
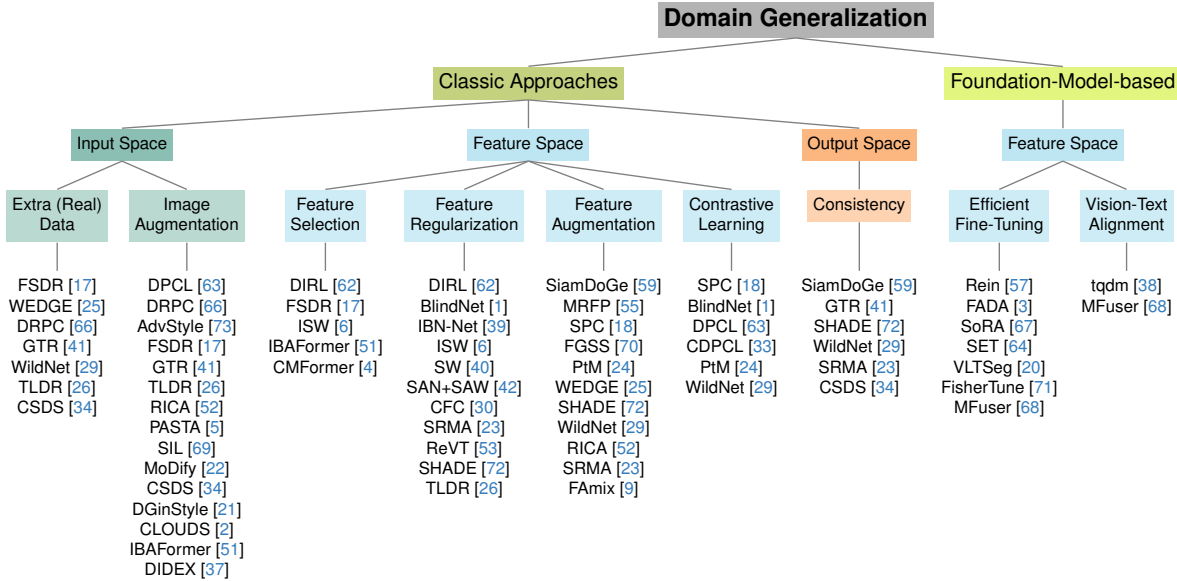
\section{Taxonomy}
Recently, a paradigm shift in DG has been observed, driven by foundation models. In contrast to "classic" DG approaches, which primarily aim at extracting generalizing features from the source domain, the new foundation-model-based methods focus on preserving and exploiting the broad and already generalizing knowledge of pre-trained models. Due to their large-scale vision or vision-language pre-training on huge web-crawled datasets, foundation models already encode highly generalized knowledge that can be adapted to the segmentation task. The labeled source dataset is then used to adapt the pre-trained model, but not primarily to learn generalizing representations. For this reason, DG approaches can first be divided into classic- and foundation-model-based approaches, as shown in the taxonomy in \Cref{fig:dg_taxonomy}.\\
Among the classic approaches, a wide variety of methods have been developed. They can be further categorized into input, feature, and output space approaches, whereas for the recently emerged foundation model approaches, only feature space methods exist so far. Several DG approaches may propose multiple contributions, for example, in both input and feature space. In this case, they appear multiple times in the taxonomy in \Cref{fig:dg_taxonomy}. Here, each subcategory, like, e.g.\ feature regularization, provides a comprehensive overview of all works. In addition to taxonomy, a comprehensive performance comparison of domain generalization approaches is provided in \Cref{tab:dg_performance} which will be discussed in \Cref{subsec:dg_performance_comparison}. 
\section{Classic Domain Generalization Approaches}
\subsection{Input Space}
\label{subsub:dg_input_space}
\textbf{Image Augmentation} Image augmentations are a simple and straightforward way to improve generalization by extending/diversifying the input distribution of the network. Therefore, in the context of DG, image augmentations are often referred to as domain randomization. Augmentations are essential for domain generalization, as they are relevant not only for input space methods, but also for, e.g. consistency losses, feature augmentations, or contrastive learning methods to obtain multiple versions of the same image.\\ 
DG approaches perform domain randomization by applying either image transformations or a style transfer, but without modifying the content. Similar to Schwonberg \etal \cite{schwonberg2023augmentation} with simple standard image augmentations, both RICA \cite{sun2023augment} and MoDify \cite{jiang2023domain} rely on image transformations. RICA \cite{sun2023augment} converts RGB images to the CIELAB color space and randomizes the mean and standard deviation along with the range of the color values, arguing that augmentations can be applied more effectively in the CIELAB color space. MoDiFy \cite{jiang2023domain} relies on a simpler randomized RGB channel swapping whose intensity is controlled by a loss memory bank. This is done in combination with a so-called difficulty-aware training strategy to adapt the difficulty of the training samples to the training progress of the network. Following the basic idea of RICA \cite{sun2023augment}, FSDR \cite{huang2021fsdr} and PASTA \cite{2023iccv_PASTA} randomize the input of the source domain in the frequency space. PASTA \cite{2023iccv_PASTA} proposes a simple augmentation by first applying a Fast Fourier Transform to the RGB image and then augmenting the amplitude spectrum. The higher the frequency, the stronger the modification. An inverse Fourier transform provides the augmented RGB image, and PASTA can be used complementary to other DG methods. FSDR \cite{huang2021fsdr} also works in the frequency domain, but accesses ImageNet samples to transfer their style to the frequency domain via histogram matching, and employs a bi-directional learning scheme to identify domain-invariant frequencies which are not randomized. As the only DG work, SIL \cite{zhang2023learning} by Zhang \etal proposes a shape augmentation to enforce shape-invariant learning. The authors use a so-called random elastic deformation algorithm to distort both the inner parts of objects and their boundaries in small local regions of the image. In addition, they use edge and contour transformations, like e.g. Canny edge detection, to extract shape information and further enforce shape-invariant learning.\\ 
Similarly to UDA, in the context of style transfer and also for feature augmentation methods (see \Cref{para:dg_feat_aug}), adaptive instance normalization (AdaIN) \cite{huang2017arbitrary} is highly relevant and widely used. AdaIN \cite{huang2017arbitrary} allows style transfer from a style reference image to a content image by transferring the feature mean and variance of the style image while preserving the content. However, unlike UDA, a target domain is not available as the style input. Several solutions have been proposed to mitigate this problem. DPCL \cite{yang2023generalized} by Yang \etal employs standard image augmentations to create different stylized images and employs AdaIN to transfer the style of the augmented images. The authors then train a network with a $L1$ reconstruction loss that reconstructs the original images from the augmented images, which should provide images similar to the source style during inference. Zhong \etal \cite{zhong2022adversarial} with AdvStyle mitigate the style references problem by formulating the mean and variance of AdaIN as learnable parameters that are learned in an adversarial manner. In a two-step process, the adversarial style-parameters are first learned and then used to augment the original source dataset. Other works use external data to randomize the style of the source domain \cite{yue2019domain, kim2023texture, liao2024class}, often with samples from ImageNet and various style transfer techniques. DRPC \cite{yue2019domain} employs an image-to-image translation model, TLDR \cite{kim2023texture} uses PhotoWCT \cite{li2018closed} for style transfer from ImageNet, and CSDS \cite{liao2024class} integrates AdaIN into its class-discriminative style transfer approach based on ImageNet. Peng \etal also use AdaIN for their GTR \cite{peng2021global} approach, but as the only DG approach, painted images are used to randomize the style of the source domain. Since the paintings contain a variety of styles, an additional module is needed to filter out too low- and too high-complexity paintings, which were found to be detrimental for style transfer.\\
In contrast to unsupervised domain adaptation, GANs are rarely used for style transfer in domain generalization approaches because the target domain is not available. However, the recently emerged text-to-image diffusion models \cite{rombach2022high} are capable of generating or translating photorealistic images without any target domain samples. DGinStyle \cite{jia2023dginstyle} and CLOUDS \cite{benigmim2024collaborating} exploit this capability to augment the original synthetic source dataset with photorealistic samples generated by a text-to-image diffusion model. CLOUDS \cite{benigmim2024collaborating} generates prompts to describe traffic scenes with an LLM and uses them to generate images with stable diffusion \cite{rombach2022high}. Since no pixel-level annotations are available for the diffusion-generated images, the authors train on these images as additional training data with pseudo-labels in a teacher-student setting. DGinStyle \cite{jia2023dginstyle} proposes a more complex framework, since it fine-tunes the diffusion models on the source domain and develops a so-called multi-resolution latent fusion module to better generate smaller objects. However, the fine-tuning step raises the problem of forgetting pre-trained knowledge, and DGinStyle proposes a three-step process including DreamBooth \cite{ruiz2023dreambooth} to preserve the original diffusion knowledge. Equipped with these two enhancements, photorealistic images conditioned on text and semantic maps are generated, and a new source domain dataset is created. It consists of half original source domain samples and half diffusion-generated samples and can be combined with any DG method. DIDEX \cite{niemeijer2024generalization} also exploits diffusion models and creates a photorealistic pseudo-target domain using image-to-image translation. Using that, it obtains a generalization by using UDA methods to adapt the model towards this pseudo-target domain.\\ 
\textbf{Extra (Real) Data} Existing methods mainly use different types of extra data: ImageNet data \cite{huang2021fsdr, yue2019domain, liao2024class, Lee2022wildnet, kim2023texture}, paintings \cite{peng2021global} and real web-crawled data such as Wedge \cite{kim2021wedge}. All methods that access additional data are listed separately in the taxonomy in \Cref{fig:dg_taxonomy}. The domain generalization paradigm assumes that there is only one (or more) labeled source domain(s), but no target domain data, which in the synthetic-to-real shift means no real-world data. However, as described above, many methods access real-world data, such as ImageNet for style transfer. The active use of real-world data gives these methods a significant advantage, because real-world styles and patterns can be incorporated into the training. While this strategy of incorporating additional real-world data is pragmatically reasonable, it compromises scientific comparability and also circumvents the real problem of domain generalization by using proxy real-world domains.  
\subsection{Feature Space}
\textbf{Feature Selection}
While all DG approaches aim at learning domain-invariant features, some approaches explicitly aim to distinguish between domain-dependent and domain-independent features in order to treat them differently. These approaches can be categorized under the term feature selection, since they treat both types of features differently. We can further distinguish between direct feature selection and indirect feature selection methods. DIRL \cite{xu2022dirl} utilizes augmentations to identify domain-invariant features. Since the augmentations introduce a style shift, the authors compute the $L2$ distance between features from the original and augmented images, and a higher distance indicates a higher domain sensitivity. This sensitivity vector guides the learning of attention weights that force the model to focus on domain-invariant features. In a very similar way with augmented images, Choi \etal \cite{choi2021robustnet} identify style-sensitive features for their instance selective whitening approach (ISW), but unlike DIRL \cite{xu2022dirl} they find these features by k-means clustering on the covariance matrix. FSDR \cite{huang2021fsdr} by Huang \etal aims to identify domain-invariant components in the frequency space. The prediction entropy serves as a criterion for identifying domain-variant frequency bands. High entropy indicates high domain dependence, and these frequencies are then randomized by histogram matching of real-world frequencies from ImageNet samples.\\
With the emergence of vision transformer architectures for domain generalization, a line of work has proposed advanced attention mechanisms to better facilitate generalized representation learning \cite{sun2023ibaformer, bi2024learning}. Since these improvements implicitly allow the networks to focus on domain-invariant features, they can be seen as indirect feature selection methods. IBAFormer \cite{sun2023ibaformer} proposes two intra-batch attention mechanisms that allow cross-image attention within a batch and should help the network to extract domain-invariant features. The mean intra-batch attention (MIBA) computes the mean of the features and allows the network attention between single images and the mean features. In contrast, the element-wise intra-batch attention (EIBA) computes the attention between all individual images of the batch separately. Finally, IBAFormer adds intra-batch attention fusion modules directly before the decoder. The authors of CMFormer \cite{bi2024learning} observed that high-resolution features are more focusing on content, while the low-resolution features are more sensitive to style. Therefore, they propose a mask-attention mechanism that processes high- and low-resolution features separately and fuses both features through a concatenation followed by a linear layer.\\
\textbf{Feature Regularization}
This category groups DG approaches that employ feature regularization objectives to constrain representation learning to domain-invariant features. Among them, several works employ instance normalization (IN) or instance whitening (IW) techniques, as these techniques have been found to be beneficial for domain generalized learning. While IN only normalizes each channel, instance whitening forces the feature covariance across channels to be decorrelated. Several works have shown that style-sensitive information are encoded in the feature covariance \cite{gatys2016image, choi2021robustnet, li2017universal} and thus decorrelation can help to learn style-invariant features.\\  
IBN-Net \cite{pan2018two} is one of the foundational works for instance normalization and batch normalization (BN). Pan \etal find that IN removes style-related information in the earlier layers of the network, thereby improving style-invariant learning, but has a detrimental effect in the deeper layers, where BN layers are beneficial to preserve content. IBN-Net is used as a complementary method in later works \cite{xu2022dirl, choi2021robustnet}. Li \etal \cite{li2022learning} argue that instance normalization can be detrimental to the spatial feature distribution. Therefore, they introduce a contour learning module to regularize the network and enforce the learning of clear object boundaries. IBN-Net \cite{pan2018two} shows that designing a network with the exact positioning of normalization layers is a challenge. For this reason, Pan \etal \cite{pan2019switchable} introduce switchable whitening (SW), which learns importance weights for either only batch and instance whitening or for five regularization techniques including batch, instance and layer normalization. In this way, the network can select the best whitening or normalization technique depending on the task and the dataset. Two other problems of whitening are the lack of awareness of already domain-invariant features and semantic classes \cite{choi2021robustnet, Peng2022semanticaware}. Choi \etal \cite{choi2021robustnet} propose instance selective whitening (ISW) based on their observation that naive whitening hinders domain-invariant discriminative features. ISW computes the variance between two covariance matrices of a non-augmented and an augmented image to identify areas sensitive to style. k-means then clusters high-variance and domain-variant features so that a filter can be applied to the naive whitening loss and only domain-variant features are decorrelated. Similarly, DIRL \cite{xu2022dirl} uses ISW, but replaces the covariance-based sensitivity feature identification with its own sensitivity vector described above and forces the 30\% most domain-variant features to be decorrelated. Peng \etal \cite{Peng2022semanticaware} develop semantic-aware normalization and whitening (SAN + SAW). Based on the classifier predictions, SAW rearranges the feature channels into groups and each group contains channels of different classes. Group instance whitening is then applied to these groups so that channels of the same class are not decorrelated, which is detrimental to the semantic features. Unlike any other whitening work, BlindNet \cite{ahn2024style} introduces a covariance regularization that enforces the covariance between the features of the original and an augmented image to be the identity matrix. In addition, Ahn \etal also introduce a direct covariance consistency by computing the covariance matrices for both images and enforcing similarity with a $L2$-loss.\\
ImageNet knowledge can be used either actively by directly accessing ImageNet samples as described in \Cref{subsub:dg_input_space}, or passively by using the encoded knowledge in pre-trained ImageNet models as a regularization. SHADE \cite{zhao2022style} and SRMA \cite{jiao2024semantic} both utilize this pre-trained knowledge of the real-world for feature space regularization. While SHADE, similar to DAFormer \cite{hoyer2022daformer}, simply uses a $L2$-loss between the frozen ImageNet features and the features of its trained model, SRMA extends this idea significantly. First, the regularization is applied to multiple layers, and second, the regularization loss has three parts: similar to SHADE \cite{zhao2022style}, an element-wise $L2$-loss, another $L2$-loss between the semantic centers of the thing classes, and a global regularization loss for the feature maps after global average pooling. TLDR \cite{kim2023texture} also implements ImageNet-based regularization, but the authors argue that direct regularization may not be optimal. For this reason, they compute the Gram matrix of both the synthetic and ImageNet features and force the Gram matrices to be similar by a $L2$-loss.\\ 
In addition to BlindNet \cite{ahn2024style}, DRPC \cite{yue2019domain} is the only work that applies a consistency loss in the feature space, which can be understood as a form of regularization. Yue \etal average the feature maps over multiple images of different styles and apply a $L1$ consistency loss that forces the features of each pyramidal feature map to be similar to the mean across all images. The approach ReVT \cite{Termoehlen2023} differs from all other feature regularization methods in that it transfers the concept of model soups \cite{wortsman2022model} to the area of domain generalization. ReVT averages the encoder weights of three encoders trained independently but with the same augmentations, and uses the new averaged encoder weights for inference. Also, averaging the decoder was found to be detrimental, so one of the three decoders is selected.\\ 
\textbf{Feature Augmentation}
\label{para:dg_feat_aug}
Similar to the input space augmentations, the feature maps of the network can also be augmented, mixed, or randomized. These techniques are closely related to the input space style transfer methods, since methods such as AdaIN \cite{huang2017arbitrary} perform style transfer or enhancement in the feature space. In addition, the mean and covariance of the features that AdaIN exploits are highly relevant, as many approaches manipulate them for feature randomization or augmentation.\\
Similar to the UDA approach DLOW \cite{gong2019dlow}, SiamDoGe \cite{wu2022siamdoge} proposes to augment the features by interpolating the feature statistics within the source domain. For this, a non-augmented and an augmented source image are processed by a Siamese-like network. Based on the resulting two feature maps, SiamDoGe randomly linearly interpolates between the original and augmented feature maps to compute new intra-domain features. SHADE \cite{zhao2022style} follows a similar idea with its style hallucination module. Starting with the base source styles, SHADE applies fast point sampling to sample styles,i.e., mean and variance. Then, similar to AdaIN, new feature statistics are computed with randomly sampled weights.\\
Style or memory banks are another common technique for feature augmentation \cite{huang2023style, zhang2023fine, kim2022pin}, and in all cases a current feature map is augmented with stored knowledge from a memory bank. FGSS \cite{zhang2023fine} implements a simple memory bank mechanism in which class-wise prototypes are stored and updated by exponential moving average. The augmented feature map is computed by a random combination of current features and the semantic class prototype. In contrast, both SPC \cite{huang2023style} and Pin the memory \cite{kim2022pin} propose a weighted feature memory bank augmentation scheme. SPC \cite{huang2023style} collects a memory bank of styles during training via EMA. For new samples during inference, the Wasserstein distance is calculated for all styles in the memory bank, and the augmented feature statistics are obtained by a weighted recombination of styles in the memory bank. The smaller the distance, the higher the weights for that style. During training, this serves as a feature augmentation with other styles, and during inference as a projection of new unknown styles into the known learned style space. In contrast, PtM \cite{kim2022pin} employs a memory matrix that is trained to contain domain-invariant features from a meta-learning framework.  The update process is performed by a small update network, and a cosine similarity matrix is computed during read access. The fused feature map is obtained as a concatenation of the original features and the weighted memory features. FAmix \cite{fahes2024simple} stands out among memory bank approaches because it is the first work to use CLIP \cite{radford2021learning}. Fahes \etal propose a two-step process in which the first step is the collection of a style bank. Unlike other works, FAmix uses prompting and the CLIP text encoder to increase the feature diversity of its style bank. The second is the training step, which uses AdaIN to linearly augment the original feature statistics with randomly sampled style bank statistics.\\
Both MRFP \cite{udupa2024mrfp} and RICA \cite{sun2023augment} employ a separate network for feature randomization. MRFP \cite{udupa2024mrfp} proposes a simple setup where a separate lightweight inverted encoder-decoder network of random convolutions randomizes the features coming from a certain layer and adds them to the output of the same layer. RICA \cite{sun2023augment} shares a similar idea, but implements a separately trained GAN directly after the first convolutional layer to augment the earlier features. Especially for deeper layers, such feature augmentation is detrimental for the generalization. The CycleGAN \cite{Zhu2017bicyclegan} architecture and principle is adopted with two generators, two discriminators, and two network streams with original and augmented input images. Hereby, the GAN effectively learns to generate features that are indistinguishable from the feature maps of the augmented network stream.\\
WEDGE \cite{kim2021wedge} and WildNet \cite{Lee2022wildnet} both propose a style transfer in the feature space. Similarly, SRMA \cite{jiao2024semantic} also uses AdaIN to normalize the features. However, instead of accessing real features, it recalculates the synthetic feature statistics by a randomly weighted linear recombination of all the individual class statistics. While both SRMA \cite{jiao2024semantic} and WildNet \cite{Lee2022wildnet} normalize the original feature statistics following AdaIN \cite{huang2017arbitrary}, WEDGE introduces a more complex style transfer into the feature space without using AdaIN. The core element of its so-called style injection is a projection matrix. First, the cosine similarity between the synthetic and real features is computed. The projection matrix then maps the synthetic features into a similar embedding space as the real features, weighted by the cosine similarity. Similarly to RICA \cite{sun2023augment}, the authors find that the style injection works best in the earlier layers and worse when applied only to the deeper layers.\\
\textbf{Contrastive Learning} Contrastive learning is a popular technique in the field of self-supervised learning and has been extensively studied for UDA \cite{Schwonberg2023Survey}. It is also widely used for domain generalization, although the construction of positive and negative pairs is more difficult due to the lack of samples from the target domain. As a substitute, DG approaches often use augmentations or style transfer to obtain style-modified image pairs for contrastive learning. Another common feature of contrastive DG approaches is the utilization of several contrastive loss functions.\\
One strategy for contrastive learning can be two different loss functions, where one loss enforces intra-cluster compactness and the other loss enforces inter-cluster dispersion of class prototypes. SPC \cite{huang2023style} and PtM \cite{kim2022pin} follow this strategy and both rely only on samples from the source domain without the need for augmentations. SPC \cite{huang2023style} proposes to perform segmentation with clustering, which requires a well-structured feature space. For this, the authors implement a clustering loss that minimizes the distance between pixel embeddings and class prototypes, and another loss adopted from the InfoNCE loss to disperse the class prototypes. The idea behind PtM \cite{kim2022pin} is the same, but with different loss functions due to the meta-learning scheme. During the meta-training phase of PtM, a memory bank is constructed and a customized compactness loss is employed to obtain features that are semantically close to the memory elements. The dispersion loss, on the other hand, ensures that the different class elements of the memory bank are well distributed in the feature space.\\
Other approaches focus on the commonly used InfoNCE \cite{gutmann2010noise} loss. Both WildNet \cite{Lee2022wildnet} and BlindNet \cite{ahn2024style} rely on similar contrastive losses and both require style-modified input pairs. WildNet \cite{Lee2022wildnet} has two InfoNCE-based losses for two differently stylized images that train the network to learn style-invariant features. They define the same pixel of their image pair as the positive pair and all pixels from other locations as the negative pairs, while excluding pixels of the same class. BlindNet \cite{ahn2024style} relies on the same definition of positive and negative pairs, but the second contrastive loss aims to learn from misclassified pixels by taking the wrong classes as negative samples. CDPCL \cite{liao2023calibration} modifies the InfoNCE loss with an uncertainty weighting matrix. This is computed from the cosine similarity between the original and augmented class prototypes and reflects how likely the class prototypes will differ between domains.\\    
Unlike other works, DPCL \cite{yang2023generalized} introduces three contrastive loss functions: pixel-pixel, class-to-pixel, and instance-to-class. Only the pixel-class loss is based on the InfoNCE loss. The pixel-pixel loss is based on transition probability matrices for both the prediction and the ground truth and is optimized by the Jensen-Shannon divergence. The ground truth transition probabilities force the features to be similar for the same class and dissimilar for other classes.    
\subsection{Output Space}
\textbf{Consistency Learning} Consistency methods rely on a straightforward principle and all have in common that the network is expected to produce the same predictions when the style or texture of the input changes. In this context, augmentations and style transfer are essential for these methods to obtain different samples with the same content, but different styles that can be used for consistency loss. GTR \cite{peng2021global} applies one of the simplest forms of a consistency loss, which is implemented as a $L1$ loss between the predictions for its locally and globally randomized input images. SiamDoGe \cite{wu2022siamdoge} follows the same principle as GTR \cite{peng2021global}, but applies a consistency loss on both output- and feature-level and weights it by a $L1$ feature distance.\\
Several other works use the KL-divergence or Jensen-Shannon divergence to enforce prediction consistency between two inputs. Both WildNet \cite{Lee2022wildnet} and CSDS \cite{liao2024class} utilize the KL-divergence. Similar to that, SHADE \cite{zhao2022style} and SRMA \cite{liao2024class} use the Jensen-Shannon divergence between different stylized images.\\
\section{Foundation-model-based Domain Generalization Approaches}
This category of DG approaches has only recently emerged, so the number and diversity of works is still limited.\\
\textbf{Efficient Fine-tuning} Similar to the basic idea of LoRA \cite{hu2021lora}, most of the approaches of this category freeze the pre-trained parameters of the foundation model and propose adapters between the frozen layers to obtain a foundation model adapted for domain generalized segmentation. Next to our own approach, Rein \cite{wei2024stronger} was one of the first approaches to follow this idea. Specifically, Rein aims to overcome the gap between the task and the pre-training by introducing a set of learnable tokens that can learn correspondences between the pre-trained features and the segmentation instances. For this, Rein computes a similarity matrix between the tokens and the features, followed by another feature fusion and an MLP. Building on this idea, both FADA \cite{bilearning} and SET \cite{yi2024learning} propose adapters in the frequency space. Their basic ideas share conceptual similarities with the classic DG approaches, such as FSDR \cite{huang2021fsdr} and PASTA \cite{2023iccv_PASTA}. FADA \cite{bilearning}, similar to many classic DG works, argues that domain-invariant information can be better learned in the frequency space and implements the Haar wavelet transform directly on the pre-trained features. The approach then splits the features into a low- and high-frequency stream, both of which have separate learnable tokens and a similarity matrix for correlation between the pre-trained features. Only for the high-frequency components, instance normalization is applied to this matrix to suppress style-dependent information. Similarly, SET \cite{yi2024learning} obtains the phase and amplitude spectra of the frozen pre-trained layers and introduces learnable tokens for both. Similar to FADA \cite{bilearning}, a similarity matrix is calculated and forwarded into a MLP. The amplitude stream, like the high-frequency stream of FADA, includes instances normalization to remove style-variant domain-dependent features. As an extension to these works, the MFuser \cite{mfuser} uses a frozen VLM and a vision-foundation model in parallel and proposes a fusion module as a connection between both networks. In contrast to the very similar, frequency-space-focused works, SoRA \cite{yun2024sora} proposes an improvement for LoRA \cite{hu2021lora} tailored to the task of domain generalized segmentation. The authors observed that only the smaller singular components carry domain-specific knowledge, while the larger components encode generalized knowledge. For this reason, SoRA fine-tunes only the minor singular components to adapt the network to the new task and preserve the pre-trained knowledge. In contrast to other approaches, VLTSeg \cite{hummer2024strong} proposes a simple setting for fine-tuning without freezing the weights and training the entire foundation model in the source domain. FisherTune \cite{zhao2025fishertune} builds on this idea and uses an approximate Fisher information matrix to estimate the domain sensitivity of the parameters. Controlled by this matrix, it only fine-tunes the most domain-sensitive parameters.\\    
\textbf{Vision-Language Alignment} The majority of works in this field do not utilize the text encoder after the vision-language pre-training, and only adopt the vision encoder for the downstream DG training. In contrast to this trend, tqdm \cite{pak2025textual} and MFuser \cite{mfuser} are the only approaches that used the text encoder during the downstream training. The authors of tqdm use the class-wise text embeddings from a frozen CLIP encoder as textual queries and text clusters for a modified cross-attention mechanism in the network decoder. The cross-attention has a text-to-pixel attention block and computes attention weights based on the text clusters, which are then used to update the pixel features. The text queries refined by the decoder are also used for the final segmentation. In addition, tqdm regularizes the network with a text, a text-vision, and a vision alignment loss. Similar to tqdm, MFuser \cite{mfuser} introduces a vision-text alignment module with an attention and a mamba block \cite{liu2024vmamba}.
\begin{table*}[]
    \centering
    \caption[Performance Comparison of domain generalization approaches]{\textbf{Performance Comparison} of all domain generalization approaches for semantic segmentation. $\csval$ denotes the validation set of Cityscapes \cite{cordts2016cityscapes}, $\bddval$ from BDD100k \cite{yu2020bdd100k} and $\mvval$ from Mapillary \cite{neuhold2017mapillary}.}
    
\extrarowheight=\aboverulesep
    \addtolength{\extrarowheight}{\belowrulesep}
    \aboverulesep=0pt
    \belowrulesep=0pt
    \renewcommand{\arraystretch}{0.7}
    \begin{tabular}{@{}lcchhhc||hhhc@{}}
        \toprule
        \textbf{DG Method} &\textbf{Enc.}&\textbf{Init}&\multicolumn{4}{c}{$\src :$ \textbf{GTA5} \cite{richter2016playing} } & \multicolumn{4}{c}{$\src :$ \textbf{SYNTHIA} \cite{ros2016synthia}} \\
        \midrule
       &&& \cellcolor{cscolor} $\csval$   & \cellcolor{bddcolor} $\bddval$ & \cellcolor{mvcolor} $\mvval$&  \raisebox{-4.0pt}{\shortstack{\textbf{DG} \\\textbf{mean}}}  \rule[-1.2ex]{0mm}{3.65ex} & \cellcolor{cscolor} $\csval$   & \cellcolor{bddcolor} $\bddval$ & \cellcolor{mvcolor} $\mvval$ &  \raisebox{-4.0pt}{\shortstack{\textbf{DG} \\\textbf{mean}}}  \rule[-1.2ex]{0mm}{3.65ex}\\
        \midrule
        
        %\parbox[t]{2mm}{\multirow{10}{*}{\rotatebox[origin=c]{90}{\small{\network{ResNet}}}}} &Baseline$^\circ$ & \ding{51} & 23.9 & 25.0 & 21.8 & 23.6\\
         %Baseline \cite{Xie2022segformer} & \parbox[t]{2mm}{\multirow{10}{*}{\rotatebox[origin=c]{90}{\small{\network{ResNet}}}}} & 36.1 & 36.6 & 43.8& 38.8\\
        DIRL \cite{xu2022dirl} &\parbox[t]{2mm}{\multirow{8}{*}{\rotatebox[origin=c]{90}{\small{\network{ResNet-50}}}}}&\parbox[t]{2mm}{\multirow{8}{*}{\rotatebox[origin=c]{90}{\small{ImageNet}}}}& 41.0 & 39.2 & 41.6 & 40.6 &- & -& - & - \\
        SiamDoGe \cite{wu2022siamdoge} &&& 43.0 & 37.5 & 40.6 & 40.4 & - & - & - & - \\
        MRPF \cite{udupa2024mrfp} &&& 42.4 & 39.6 & 44.9 & 42.3 & 27.9* & 25.7* & 26.0* & 26.5* \\
        SPC \cite{huang2023style} &&& 44.1 & 40.5 & 45.5 & 43.4 &- & - & -& - \\
        BlindNet \cite{ahn2024style} &&& \textbf{45.7} & \textbf{41.3} & \textbf{47.1} & \textbf{44.7} & -&- &-&- \\
        FGSS \cite{zhang2023fine} &&& 44.5 & 36.5 & 44.3 & 41.8 & 39.2 & 29.8 & 33.5 & 34.2 \\
        DPCL \cite{yang2023generalized} &&& 44.9 & 40.2 & 46.7 & 43.9 & - & -& -& - \\
        CDPCL \cite{liao2023calibration} &&& 42.3 & 39.5 & 42.4 & 41.4 & \textbf{41.2} & \textbf{35.4} & \textbf{35.5} & \textbf{37.4} \\
        \hline
        IBN-Net$^\circ$~\cite{pan2018two}&\parbox[t]{2mm}{\multirow{18}{*}{\rotatebox[origin=c]{90}{\small{\network{ResNet-101}}}}} &\parbox[t]{2mm}{\multirow{18}{*}{\rotatebox[origin=c]{90}{\small{ImageNet}}}}  & 37.7 & 34.2 & 36.8 & 36.2 & -& -& -& -\\
        ISW$^\circ$~\cite{choi2021robustnet}&&&  37.3 & 38.7 & 38.1 &38.0 &- & -& -& -\\
        DRPC~\cite{yue2019domain}&&&  42.5& 38.7& 38.1&39.8&37.6&34.3&34.1&35.4\\
        SW~\cite{pan2019switchable}& && 36.1 &36.6&32.6&35.1 &31.6 & 35.5 & 29.3 & 32.1 \\
        AdvStyle \cite{zhong2022adversarial} &&& 39.5 & 36.4 & 36.1 & 37.3 & 37.6 & 27.5 & 31.8 & 32.3 \\ 
        FSDR~\cite{huang2021fsdr}& && 44.8 &41.2&43.4&43.1 &40.8&39.6&37.4&39.3 \\
        PtM \cite{kim2022pin} &&& 44.9 & 39.7 & 41.3 & 42.0 & - & -& -& - \\ 
        SAN+SAW~\cite{Peng2022semanticaware}& && 45.3 &41.2& 40.8 &42.4 &40.9&36.0&37.3&38.1\\
       WEDGE \cite{kim2021wedge}&&& 45.2&41.1&48.1&44.8 & 40.9 & 38.1 & \textbf{43.1} & 40.7 \\
       GTR \cite{peng2021global}&&&43.7 &39.6&39.1&40.8 & 39.7 & 35.3 & 36.4 & 37.1 \\
        SHADE \cite{zhao2022style}&&&46.7 &43.7&45.5&45.3 & - & -& -& -\\
        WildNet$^\circ$~\cite{Lee2022wildnet}  & && 45.8 & 41.7 & 47.1 &44.9 & -& -& -& -\\
        CFC \cite{li2022learning} &&& 41.1 & 37.8 & 44.4 & 41.1 & - & - & - & -\\
        TLDR~\cite{kim2023texture}  &&&  {47.6} & {44.9} & {48.8} & {47.1} & 42.6 & 35.5 & 37.5 & 38.5\\
        RICA~\cite{sun2023augment}  & && {48.0} & 45.2 & {46.3} & {46.5} & \textbf{45.0} & 36.3 & 41.6 & 41.0\\
        PASTA~\cite{2023iccv_PASTA}  & && {45.3} & 42.3 & {48.6} & {45.4} & -& -& -& -\\
        SIL \cite{zhang2023learning} &&& 48.0 & 40.3 & 47.4 & 45.2 & 40.9 & 33.4 & 33.4 & 35.9 \\
        MoDify \cite{jiang2023domain} &&& \textbf{48.8} & 44.2 & 47.5 & 46.8 & 43.4 & \textbf{39.5} & 42.3 & \textbf{41.7} \\
        SRMA \cite{jiao2024semantic} &&&48.4 & 45.0 & 49.6 & 47.7 & - & -&- & -\\
        CSDS \cite{liao2024class} &&&48.5 & \textbf{46.1} & 49.4 & \textbf{48.0} & 43.9 & 38.2 & 40.6 & 40.9 \\
        DGinStyle \cite{jia2023dginstyle} &&& 46.9 & 42.8 & \textbf{50.2} & 46.6 & -& -& -& - \\
        \hline
        FAMix \cite{fahes2024simple} &\parbox[t]{2mm}{\multirow{2}{*}{\rotatebox[origin=c]{90}{\small{\network{R101}}}}}&\parbox[t]{2mm}{\multirow{2}{*}{\rotatebox[origin=c]{90}{\small{CLIP}}}}& 49.5 & 46.4 & 52.0 & 49.3 & -& -& -& - \\
        CLOUDS \cite{benigmim2024collaborating} &&& \textbf{55.7} & \textbf{49.3} &  \textbf{59.0} & \textbf{54.7} & \textbf{49.1} & \textbf{40.3} & \textbf{50.1} & \textbf{46.5}\\
         %DIDEX \cite{niemeijer2024generalization} &&& 52.4 & 40.9 & 49.2 & 47.5\\
         \hline
         HRDA \cite{hoyer2023domain} &\parbox[t]{2mm}{\multirow{5}{*}{\rotatebox[origin=c]{90}{\small{\network{MiT-B5}}}}}&\parbox[t]{2mm}{\multirow{5}{*}{\rotatebox[origin=c]{90}{\small{ImageNet}}}}& 57.4 & 49.1 & 61.2 & 55.9 & -& -& -& -\\ 
         DGinStyle \cite{jia2023dginstyle} &&& 58.6 & {52.3} & {62.5} & {57.8} & -& -& -& - \\
         DIDEX \cite{niemeijer2024generalization} &&& \textbf{62.0} & \textbf{54.3} & \textbf{63.0} & \textbf{59.7} & \textbf{59.8} & \textbf{47.4} & \textbf{59.5} & \textbf{55.6}\\
         ReVT \cite{Termoehlen2023} &&& 50.0 &  48.0 & 52.8 & 50.2 & 46.3 & 40.3 & 44.8 & 43.8 \\
         IBAFormer \cite{sun2023ibaformer} &&&56.3 & 49.8 & 58.3 & 54.8 & {50.9} & {44.7} & {50.9} & {48.7} \\
         \hline
         CMFormer \cite{bi2024learning} & Swin-L & IN1k &  55.3 & 49.9 & 60.1 & 55.1 & 44.6 & 33.4 & 43.3 & 40.4 \\
         VLTSeg \cite{hummer2024strong} & EVA02-L & EVA02-CLIP & 65.3 & 58.3 & 66.0 & 63.2 & 56.8 & 51.9 & \textbf{55.1} & 54.6 \\
         tqdm \cite{pak2025textual} & EVA02-L & EVA02-CLIP & 68.9 & 59.2 & 70.1 & 66.1 & \textbf{58.0} & \textbf{52.4} & {54.9} & \textbf{55.1} \\
         MFuser \cite{mfuser} & EVA02-L & EVA02-CLIP & \textbf{70.2} & \textbf{63.1} & \textbf{70.1} & \textbf{71.3} & 54.2 & 46.7 & 53.2 & {51.4} \\
         \hline 
         Rein \cite{wei2024stronger} &\parbox[t]{2mm}{\multirow{5}{*}{\rotatebox[origin=c]{90}{\small{\network{ViT-L}}}}} &\parbox[t]{2mm}{\multirow{5}{*}{\rotatebox[origin=c]{90}{\small{DinoV2}}}} & 66.4 & 60.4 & 66.1 & 64.3 & -& - & -& -\\
         SET \cite{yi2024learning} &&& 68.1 & 61.6 & 67.7 & 65.8 & 49.7 & 45.5 & 49.5 & 48.2\\ 
         SoRA \cite{yun2024sora} &&& \textbf{71.8} & 61.3 & \textbf{71.7} & \textbf{68.3} & - & -& -& -\\
         FADA \cite{bilearning} &&& 68.2 & {61.9} & 68.1 & 66.1 & \textbf{50.0} & \textbf{45.8}& \textbf{49.9} & \textbf{48.6} \\
        FisherTune \cite{zhao2025fishertune} &&& 68.2 & \textbf{63.3} & 68.0 & 66.5 & - & - & - & - \\

        \bottomrule
    \end{tabular}

    \label{tab:dg_performance}
    %\vspace{1mm}
\end{table*}
\section{Performance Comparison}
\label{subsec:dg_performance_comparison}
Similar to UDA, GTA5 \cite{richter2016playing} and Synthia \cite{ros2016synthia} are used as synthetic source datasets for the vast majority of DG approaches. Cityscapes \cite{cordts2016cityscapes}, BDD100k \cite{yu2020bdd100k} and Mapillary \cite{neuhold2017mapillary} are used as the unseen datasets for DG performance evaluation. This setting allows for a very consistent and comparable benchmarking of DG approaches. For this reason, a list of the performance of all DG approaches is provided for both synthetic source datasets in \Cref{tab:dg_performance}.\\
First, we can clearly observe how the different backbone architectures \network{ResNet-101}, \network{MiT-B5} and \network{ViTs} and their pre-trainings affect the domain generalization capabilities. While the best performing approach with a \network{ResNet-101} architecture and ImageNet initialization reaches 48.0\% mIoU for the DG Mean, this increases to 54.7\% mIoU for CLOUDS \cite{benigmim2024collaborating} with a CLIP initialization while the approach clearly benefits from three large foundation models. However, when the SegFormer \cite{xie2021segformer} backbone \network{MiT-B5} is used, the current peak performance reaches 57.8 \% mIoU. Finally, the foundation models with their large-scale pre-training raise the DG performance to another level, and the very recent approach SoRA \cite{yun2024sora} achieves 68.3 \% mIoU for the average over three real-world datasets.\\
The list of DG approaches in \Cref{tab:dg_performance} also confirms the observation from the UDA analysis that generalizing from the Synthia dataset is more difficult than for GTA5. For example, if we look at the absolute peak performance across all architectures and initializations, we see that it is significantly lower at 55.1\% mIoU compared to GTA5 at 68.3\%. Notably, the performance for GTA5 is reported across 19 classes and for Synthia as the mean of 16 classes. The newer foundation model approaches also clearly struggle when trained on Synthia. While the smaller dataset size (9.400 for Synthia vs. 24.499 for GTA5) is given as a reason \cite{bi2024learning} for the lower performance, other reasons such as the different viewpoint of Synthia and also the lower rendering quality should be taken into account.\\
Another trend can be observed in the given performance comparison. It is easy to see that the performance gains of new DG approaches with a \network{ResNet-101} backbone are comparatively small. Between TLDR \cite{kim2023texture} from early 2023 and the current leading approach CSDS \cite{liao2024class}, there is a small improvement of only 0.9\% mIoU in the DG mean. Also, the other performances show that many approaches achieve a performance in a similar range without a major gain. This trend is similar to UDA, where stagnation in performance improvements was also observed. However, similar to the field of unsupervised domain adaptation, new architectures like the \network{MiT-B5} and foundation models \cite{wei2024stronger, hummer2024strong, yun2024sora} have enabled new breakthroughs in performance and methodology.
\section{Conclusion \& Outlook}
Overall, this survey provides a comprehensive overview of the quickly evolving and relevant field of domain generalization for semantic segmentation. We identified a paradigm change from classic DG approaches towards foundation-model approaches which has enabled significant performance progress. For both classic and foundation-model-based approaches, we define a fine-grained clustering of input, feature, and output space approaches. Domain generalization with foundation models is a promising future research direction because the research in this area is still at an initial stage but shows a strong performance. Beyond this, it can also be valuable to extend the scope of DG research towards end-to-end automated driving approaches and research domain generalization on a system-level. 
\vspace{-2mm}
\section*{Acknowledgment}
This work was supported by the German Federal Ministry for Economic Affairs and Climate Action within the project “Safe AI Engineering” under Grant 19A24004U.
{
    \small
    \bibliographystyle{ieeenat_fullname}
    \bibliography{main}
}


\end{document}